\documentclass[11pt]{article}
\usepackage{coling2020}
\usepackage{times}
\usepackage{url}
\usepackage{latexsym}

\usepackage{multirow}
\usepackage{graphicx}
\usepackage{amsmath}
\usepackage{url}

\usepackage{xcolor}
\usepackage{subcaption}

\definecolor{comment}{RGB}{255, 0, 0}

\colingfinalcopy 

\title{Enhancing Product Safety in E-Commerce with NLP}

\author{Kishaloy Halder$^{\dagger}\thanks{~~Work done prior to joining Amazon}$ \hspace{0.5cm} Josip Krapac \hspace{0.5cm} Dmitry Goryunov$^{\dagger}$   \hspace{0.5cm}  Anthony Brew$^{\dagger}$ \\
{\bf Matti Lyra} \hspace{0.5cm} {\bf Alsida Dizdari$^{\dagger}$} \hspace{0.5cm} {\bf William Gillett} \hspace{0.5cm} {\bf Adrien Renahy} \hspace{0.5cm} {\bf Sinan Tang} \\\\
  $^\dagger$ Work done while at Zalando \\\\
  {\small \texttt {kishaloh@amazon.com \hspace{0.2cm} josip.krapac@zalando.de \hspace{0.2cm} d.f.goryunov@gmail.com}}\\
  {\small \texttt {atbrew@gmail.com \hspace{0.2cm} matti.lyra@zalando.de \hspace{0.2cm} alsida.dizdari@gmail.com}} \\
  {\small \texttt {william.gillett@zalando.de \hspace{0.2cm} adrien@zalando.fr \hspace{0.2cm} sinan.tang@zalando.de}} \\
  }

\date{}

\begin{document}
\maketitle
\begin{abstract}
  Ensuring safety of the products offered to the customers is of paramount importance to any e-commerce platform. Despite stringent quality and safety checking of products listed on these platforms, occasionally customers might receive a product that can pose a safety issue arising out of its use. In this paper, we present an innovative mechanism of how a large scale multinational e-commerce platform, Zalando, uses Natural Language Processing techniques to assist timely investigation of the potentially unsafe products mined directly from customer written claims in unstructured plain text. We systematically describe the types of safety issues that concern Zalando customers. We demonstrate how we map this core business problem into a supervised text classification problem with highly imbalanced, noisy, multilingual data in a AI-in-the-loop setup with a focus on Key Performance Indicator (KPI) driven evaluation. Finally, we present detailed ablation studies to show a comprehensive comparison between different classification techniques. We conclude the work with how this NLP model was deployed.
\end{abstract}

\section{Introduction}
\label{section:intro}
Keeping the platform safe for all customers is one of the top priorities for many (if not all) large e-commerce businesses \cite{ullrich2019new,satheeshkumar2021effective}. As a popular e-commerce platform for fashion (clothes, shoes, accessories) and beauty products (cosmetics), Zalando\footnote{\url{zalando.de}} observes a large number of customer returns for size and fit, or quality related reasons. While such returns are not surprising as fashion is often manifested from personal preferences, occasionally customers report safety issues with products such as broken heel or sharp protruding edges. We consider such occurrences as Product Safety (PS) cases in this work.
To investigate the gravity of individual cases with due diligence, the customers are required to submit a description of their experience with the product in plain text, along with an optional image of the product while reporting such a case.

\begin{figure}[t]
    \centering
    \includegraphics[height=3in]{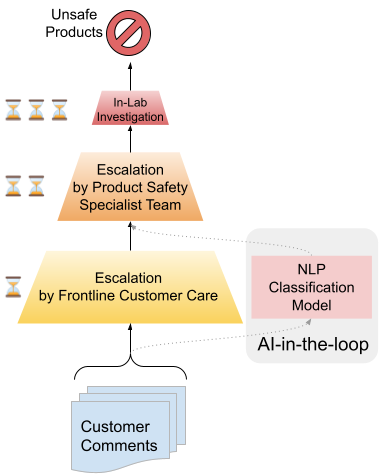}
    \caption{Product Safety investigation workflow in Zalando. A Legacy, end-to-end manual Workflow is shown on the LHS. The proposed and newly implemented AI-in-the-loop workflow is on the RHS.
    }
    \label{fig:ps_workflow}
\end{figure}

We break down the overall PS investigation process in three 
stages as depicted in Figure \ref{fig:ps_workflow}. The process starts from the customer submitted comments. In the first stage, all
the customer claims are examined by the Front-line Customer Care (FCC) agents. They escalate some of these cases as 
potential PS issues which are usually rare and thus much smaller in volume compared to that of all the customer comments. In the next stage of investigation, a team of PS agents (PS team hereafter) who have gone through specialized training to spot PS related issues from the customer comments, examine the escalated cases. They further filter escalated cases after close investigation. A tiny fraction of the escalated cases, which they perceive to be truly related to safety, are forwarded to the next stage of the process \textit{i.e.,} laboratory investigation. 
Based on the findings the PS team takes necessary actions against the unsafe products \textit{e.g.,} take down product listing and escalate the case to the manufacturer. Finally, they also record the outcome of the investigations and whether each case is indeed about safety. 

As the escalations traverse through the stages, the depth of scrutiny of individual cases increases substantially, and so does the operational cost. To make the optimal use of the investigative resources, it is imperative that at each stage, only those cases with high likelihood to be PS related, are escalated.

The rarity of true PS issues creates a unique challenge in keeping the platform \textit{safe}. A diverse set of issues can be misplaced as being generic quality issues by the regular FCC agents, risking thus missing true PS cases.

In this work, we present a Natural Language Processing (NLP) based framework to address this core problem of flagging \textit{all} potential Product Safety cases without overwhelming the limited capacity of PS experts with irrelevant cases. To summarize our contributions are the following:

\begin{itemize}
    \item We present a systematic overview for the problem of PS investigation in a large-scale ecommerce company.
    \item We identify key signals in the investigation workflow to map this business problem into a data-driven modeling task.
    \item We demonstrate how modern NLP techniques can be used in conjunction with strategic training procedures to overcome the challenges in a production such as class imbalance, noise, and multilinguality.
\end{itemize}
\section{Problem Description}
\label{section:problem_description}
PS cases are a subset of all customer claims where the comment indicates a broader safety issue from using the product. The PS team (second stage in Figure \ref{fig:ps_workflow}) categorizes all such claims into the four categories presented in Table \ref{tab:case-definition}.

\begin{table}[]
\centering
\resizebox{0.6\textwidth}{!}{%
\begin{tabular}{l|l}
\hline
\textbf{Case Type} & \textbf{Criterion} \\ \hline
Allergic Reaction & \begin{tabular}[c]{@{}l@{}}Rashes/Redness, Itching/Irritation\\ Pimples, Burning Sensation \\Swollen skin, Shortness of breath \dots\end{tabular} \\ \hline
Chemical Smell & \begin{tabular}[c]{@{}l@{}}Fish smell\\ Petrol/gasoline/diesel smell\\ Sulphur/Chlorine smell\\ Strong dye smell \dots\end{tabular} \\ \hline
Injury & \begin{tabular}[c]{@{}l@{}}Nail sticking out\\ Needle/Pins/Metal object found\\ Sharp/rough edge\\ Slippery shoes, Broken heel \dots\end{tabular} \\ \hline
Not Product Safety & \begin{tabular}[c]{@{}l@{}}Perfume smell, Sweat smell\\ Torn label, Signs of wear\\ Cigarette/smoke smell, Blisters\\ Stains, creases, scratches \dots\end{tabular} \\ \hline
\end{tabular}%
}
\caption{Overview of (a non-exhaustive) criterion for Product Safety cases addressed in Zalando. Complaints mentioning ``Allergic Reaction'', ``Chemical Smell'', ``Injury'' are regarded as potentially unsafe and are duly investigated. 
}
\label{tab:case-definition}
\end{table}

In this work we regard the classifications obtained from the PS team as the gold-standard, ground-truth data. The first three case types \textit{i.e.,} ``Allergic Reaction'', ``Chemical Smell'', and ``Injury'' are formally termed as PS cases, and are further investigated following lab-protocols. A significant section of all the comments is regarding generic quality-related issues, and is marked as ``Not Product Safety'' by the PS specialist team.

\subsection{Practical Challenges}
\label{subsec:challenges}
This core business problem of identifying PS cases from the customer claims early comes with a unique set of challenges, especially when applying the process to the scale of Zalando's large customer base (46.3 million active customers in as of Q3 2021). \\ 

\noindent \textbf{1. Multilinguality:} Currently Zalando operates in 23 countries where 18 different languages are spoken: German (Germany, Austria, Switzerland), Polish, French (France, Switzerland, Belgium), Dutch (the Netherlands, Belgium), English (UK, Ireland), Italian (Italy, Switzerland), Spanish, Swedish, Danish, Norwegian, Finnish, Czech, Slovak, Estonian, Latvian, Lithuanian, Slovenian, and Croatian. While it is important that the PS cases are identified with due diligence irrespective of the languages the customers speak, this ever-expanding\footnote{Zalando launched its services in $6$ language markets in Q2, 2021} list of languages makes classification difficult. Additional challenge is that the annotated ground-truth data majorly consists of German ($90\%$), Dutch ($3\%$), and Polish ($3\%$) comments. For many of the languages mentioned above, there is no annotated comments available at all.\\ 

\noindent \textbf{2. Label Distribution Mismatch:} The ground-truth labels obtained by the PS team are assigned only to the cases escalated by the FCC. Therefore we observe a much larger percentage of PS cases in the ground-truth data ($40\%$) compared to the set of all comments that enter the system ($<1\%$) for PS cases. In other words, the label distributions between the training and inference are drastically different. This mismatch presents a challenge for any classification system to work reliably in production traffic.
\section{Problem Formulation}
\label{section:problem_formulation}

We formulate the detection of PS cases as a binary text classification problem. Each customer comment falls into one of the two classes: ``Product Safety'' or ``Not Product Safety''. This binary (rather than a four-way) formulation of the classification problem is a conscious choice since, (i) the PS team investigates a case as long as it is \textit{any} of the three PS classes, (ii) the binary objective leaves less room for confusion to an underlying classifier compared to the four-way alternative. Hereafter, we refer to ``Product Safety'' as positive class, and ``Not Product Safety'' as negative class interchangeably. \vspace{-0.2cm}

\subsection{Methodology}
\label{subsec:methods}
As presented in Figure \ref{fig:ps_workflow}, we augmented the front-line escalation process with an AI-in-the-loop system to help safeguard the safety of customers. We employ a text classification model trained on the annotations obtained from the PS team in parallel with the manual escalation by the FCC agents.\\ \vspace{-0.2cm}

\noindent \textbf{Model Architecture:} To build the classifier, we employ a BERT model \cite{devlin2019bert} pre-trained on Wikipedia corpus from the top 104 languages\footnote{https://huggingface.co/bert-base-multilingual-cased}. It uses a shared WordPiece vocabulary built from all the
languages to tokenize input texts. The top $104$ languages cover all the languages that concern Zalando, as such 
we do not pre-train the model with a Masked Language Modelling objective on any additional corpus. The tokenizer converts the sequence of words into WordPiece token-ids. The BERT model takes this sequence as input, and the representation
of the {\tt [CLS]} token from the last layer of the BERT network is considered as the finite dimensional ($768$) representation of the
text. A linear layer is used to project that onto a $2$-dimensional vector, and {\tt softmax} activation 
is used to transform it into likelihood scores between two classes. We use the Adam optimizer\cite{kingma2015adam} to minimize {\tt cross-entropy} loss, both learning the parameters of the linear layer and fine-tuning the BERT model. \\ \vspace{-0.2cm}

\noindent \textbf{Addressing Multilinguality:} 
It is imperative that we only deploy a model to production traffic when its performance on unseen, multilingual data is not only acceptable, but also can be reliably estimated. In the absence of any data many languages, there are two options to address the multilinguality aspect, \textit{i.e.,} (i) always use a translated version of a comment into a common language (\textit{e.g.,} English) during model training, as well as inference; (ii) translate a portion of the available corpus to other languages with less or \textit{zero} comment share, and augment the dataset. 

The first option simplifies the modelling process as all the comments are always in the same language. It also facilitates understanding of the content for PS team experts as translation brings down the language barrier. However, this flexibility comes at the cost of translating a large amount of comments that enter the system (with $\sim99\%$ of them not being PS anyway). On the other hand, the second option of using a multilingual augmented corpus eliminates the need
for a such large number of on-the-fly translations, as the trained model can classify comments irrespective of the language, thus bringing operational costs down. Therefore, we choose the second option \textit{i.e.,} multilingual corpus augmentation. We translate all the comments in the annotated corpus to the $17$ other languages, and augment the corpus with these
translated comments along with the same associated labels (assuming the classification of a comment does not change if the same is expressed in another language).\\

\begin{figure}[t]
    \centering
    \includegraphics[height=2.3in]{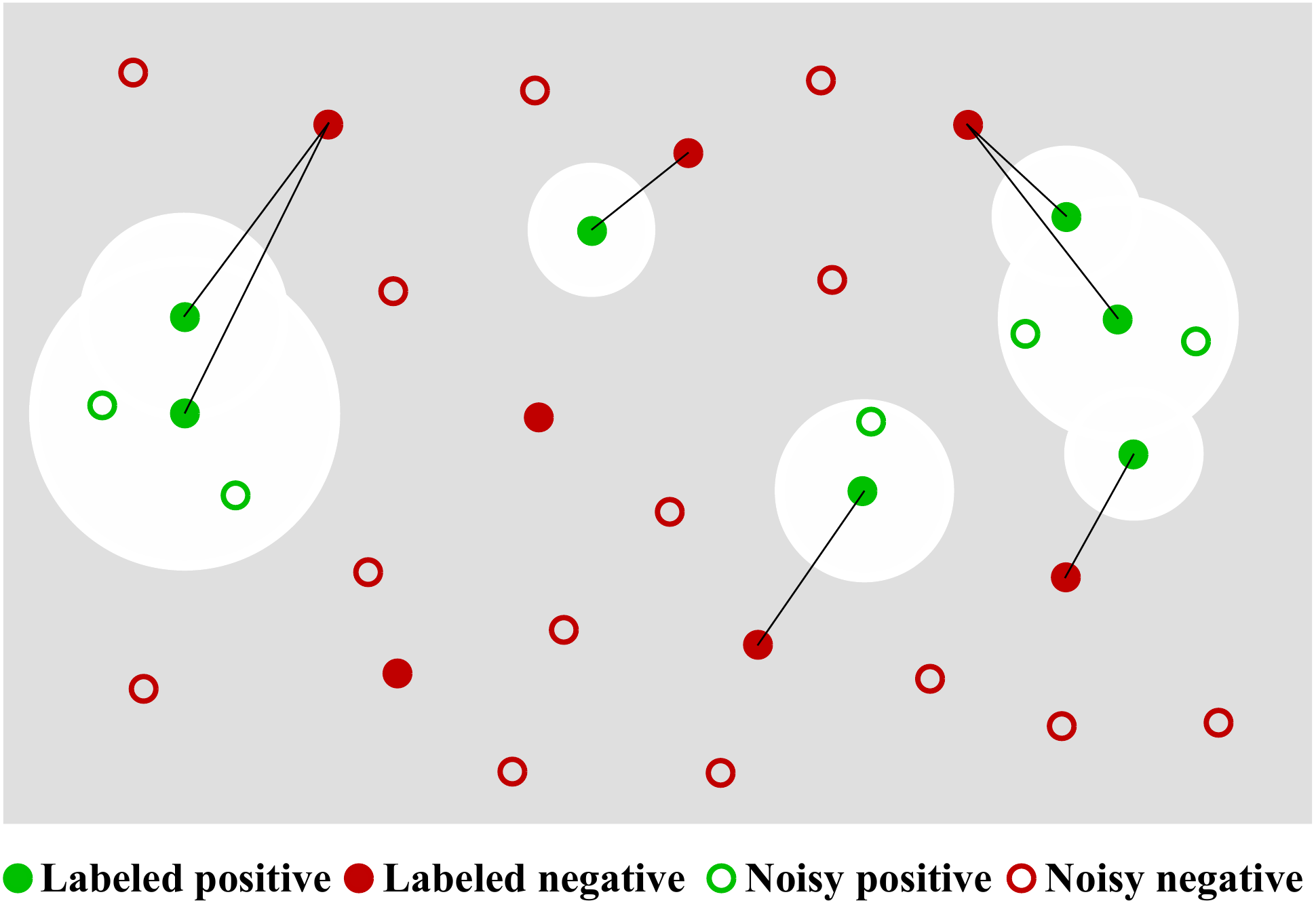}
    \caption{Illustration of Noisy Negative mining. For each labeled positive, we calculate the distance to the closest labeled negative, and multiply it with a factor $\beta\in[0,1]$ ($0.5$ in the figure) to determine the radius of a ball (circle in 2D) centered at each labeled positive point. Any unlabeled point outside of the union of the positive balls is noisy negative.
    }
    \label{fig:noisy_neg_mining}
\end{figure}

\noindent \textbf{Addressing Label Distribution Mismatch:}
A common practice to address this issue is to bias the class-weights or under/over sample labelled instances during model training. We adopt a different route in this regard. Note that in our setting, there is limited availability of gold-standard annotated data, however, there is an abundant pool of comments that are filtered out by the FCC on a daily basis for not being related to PS. We mine a proportionate chunk of \textit{noisy} negative comments from this pool of filtered out comments. We use a multi-lingual {\tt Sentence-Bert} model to encode all the comments \cite{reimers-2019-sentence-bert}. We ensure the mined negatives are not similar to any of the PS comments as shown in Figure \ref{fig:noisy_neg_mining}. 

\section{Experiments}
\label{sec:experiments}
Since our text classification system needs to handle production traffic, we use Key Performance Indicators (KPI) to evaluate different methodologies. The KPIs are constrained by business requirements:

\noindent{\textbf{1.} The recall of the system has to be as high as possible to ensure the likelihood of missing a true PS case is minimal.}\\ 
\noindent{\textbf{2.} The precision of the system has to be high to ensure the manageable workload for the PS team.}

\begin{table}[ht]
\centering
\resizebox{0.8\textwidth}{!}{%

\begin{tabular}{lcccc} \hline
\multirow{2}{*}{Dataset} & Train & Dev & Test & Traffic \\
 & size / \#words & size / \#words & size / \#words & size / \#words \\ \hline
Original & 12.7K / 42.62 & 1.3K / 35.98 & \multirow{3}{*}{11,2K / 32.20} & \multirow{3}{*}{138,3K / 30.22} \\
Original+NN & 281K / 29.20 & 32.3K / 26.68 &  &  \\
Original+NN+PC & 418K / 34.60 & 46K / 29.93 &  & \\ \hline
\end{tabular}%
}
\caption{Dataset Statistics.}
\label{tab:data-stats}
\end{table}

\subsection{Definition of Data Splits}
\label{subsec:data_splits}
We define four data splits (\textit{i.e.,} train, dev, test, and traffic) to train and evaluate our model considering the live production scenario. This is a live system and the nature of the comments might evolve,
so we define the splits according to the comments' generation time. Let us assume we have annotated data since $t=t_0$ and till $t=t_N$ denoting data from $0^{\text{th}}$ to $N^{\text{th}}$ (\textit{i.e.,} current) month. We hold out last month's data \textit{i.e.,} $t_{N-1}$ to $t_N$ as the test data. This is a conscious decision to evaluate the model on the most recent ground-truth data to get a reliable estimate of how it might perform on the near-future production traffic. We divide the rest of the ground-truth data (\textit{i.e.,} $t_0$ to $t_{N-1}$) in a $90:10$ proportion between train, and dev data. Finally, we consider all the comments in production traffic that are generated between $t_{N-1}$ and $t_N$ as the traffic data.

Note that the test is just a tiny, labelled subset of the traffic split (\textit{cf} Figure \ref{fig:ps_workflow}). To perform a fair comparison between different models, the test and traffic data are kept identical for different models. The statistics are presented in Table \ref{tab:data-stats}. We experiment with three forms of the training data -- Original: without any augmentation; Original+NN: augmented with the Noisy Negatives mined from unlabelled data; Original+NN+PC: augmented with both Noisy Negatives as well as the Parallel Corpus with the translated comments.

\subsection{Evaluation}
\label{subsec:evaluation}
Note that, the standard classification metrics such as precision and recall can only be reported on the test. To measure the adherence to the second business requirement, we report the volume of comments classified as positive out of all comments in traffic data (denoted by Model $\cup$ FCC). We fine-tune the BERT models separately for each of the three datasets mentioned above using \textsc{Flair}\cite{akbik2019flair}. We used the following hyperparameters, mini batch-size $32$, learning rate $3e^{-6}$, and early stopping based on dev loss. In order to maximize the call as per the first business requirement, we estimate a prediction threshold based on $95\%$ recall on the corresponding dev set for all the models.

Table \ref{tab:kpi_scores} shows that the model trained on the Original dataset, due to mismatched class prior, results in more than $16K$ forwarded cases in a month. The model trained on Original+NN performs better as it has access to the Noisy Negatives, resulting in a volume of $3.4K$. The model trained on our proposed Original+NN+PC dataset achieves the best recall of $92\%$ on the test data, as well as it attains the lowest volume of $2.7K$. \\

\noindent{\textbf{Language Fairness:}} In the rightmost column (Avg. Std.), we show how \textit{fairly} each model treats comments in different languages. For every comment in the test data, we also have the translated versions in the other 
languages. Ideally, a model should infer identical likelihood scores for a comment
irrespective of the language it is written in, yielding a standard deviation of $0$ across different language versions of the same comment. We compute the average of such standard deviations across all samples in the test, where lower scores are better. We observe that despite using multi-lingual BERT as the starting point, the model trained on Original dataset has the highest average standard deviation. Our proposed Original+NN+PC results in the lowest standard deviation in this regard, indicating that parallel corpus training is needed to treat languages fairly. 

\begin{table}[]
\centering
\resizebox{0.8\textwidth}{!}{%
\begin{tabular}{lccccc}
\hline
\begin{tabular}[l]{@{}l@{}}Model\\Training Dataset\end{tabular} & Precision & Recall & \begin{tabular}[c]{@{}c@{}}Volume\\ Model $\cup$ FCC\end{tabular} & \begin{tabular}[c]{@{}c@{}}Volume\\ Model\end{tabular} & \begin{tabular}[c]{@{}c@{}}Avg.\\ Std.\end{tabular} \\ \hline
Original & $0.60$ & $0.78$ & $16,165$ & $16,102$ & $0.20$ \\
Original+NN & $\mathbf{0.75}$ & $0.52$ & $3416$ & $3342$ & $0.15$ \\
Original+NN+PC & $0.63$ & $\mathbf{0.92}$ & $\mathbf{2782}$ & $\mathbf{2714}$ & $\mathbf{0.12}$ \\ \hline
\end{tabular}%
}
\caption{Experimental results with different training methodologies. Our proposed training method achieves the highest Recall, and lowest Volume of escalated comments.}
\label{tab:kpi_scores}
\end{table}
\section{Production}
\label{section:production}

We describe the implementation of the training, and deployment pipelines and demonstrate how we address deployment challenges and business requirements specific to the core problem.\\ \vspace{-0.2cm}

\begin{figure}[t]
    \centering
    \includegraphics[height=2in]{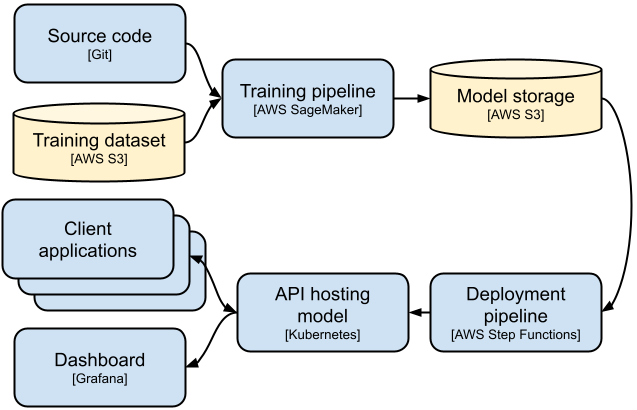}
    \caption{Production deployment of the model.}
    \label{fig:deployment}
\end{figure}


\noindent \textbf{Training Pipeline:} 
Reproducibility of models is a key aspect to maintain and compare baselines in production. To ensure that, we keep all the code related to data preparation and model training in a Git repository which is version controlled by design. Any change to the training code is reflected in the repository before the model gets trained, thus allowing reproducing the result of previous models.
As part of GDPR \cite{voigt2017eu}, Zalando deletes customer data on request, which might cause in data discrepancy between runs that are performed with a significant time gap. 
We address this by ensuring that a newly trained candidate model and the baselines are always compared on the identical evluation data.
To run the training pipeline, we use AWS SageMaker\footnote{\url{https://aws.amazon.com/sagemaker/train/}} which can be spawned on demand and bills for expensive resources (\textit{e.g.,} GPUs) only for the duration of the model training.\\


\noindent \textbf{Deployment Pipeline:} 
It is important to version each model trained and deployed in production for auditing performances of historical models. We achieve that by time-stamping every prediction and version controlling all the deployed models. This practice ensures that we can always link a prediction back to a model version. All model versions are stored on AWS S3 enabling comparison of a new candidate model with earlier ones.
When we release a model into production, the deployment pipeline downloads the model binary from AWS S3 and loads it into the memory of an API deployed on Kubernetes\footnote{\url{https://kubernetes.io/}}.
It allows us to scale the service and enables blue-green and canary deployment \cite{bluegreen2018}. These techniques help in reducing the downtime of this critical service.\\


\noindent \textbf{Performance Monitoring:} We develop a dashboard reporting the performance of the production model in various time windows (\textit{e.g.,} days, weeks, months). Based on business priorities, the dashboard displays per-language Precision, Recall and the Volume for Model $\cup$ FCC. The dashboard allows us to measure if a candidate model outperforms the production one, before the actual deployment. Additionally, discrepancies between performance on test data and dashboard help in detecting issues in data pipelines.
\section{Conclusion}
\label{section:conclusion}
In this work, we address the problem of identifying Product Safety related issues in Zalando offerings. We presented different categories of safety issues and mapped this business problem into a binary text classification task where the input modality is customer written feedback. We tackle two major challenges: multilingual input and train-inference label distribution mismatch. We propose to use multilingual BERT model with translated data to address the first challenge, and use data augmentation technique based on mining noisy negatives from unlabelled data to address the second. 

The new fully compliant system now classifies incoming customer claims in real time in the $23$ Zalando markets. It ensures customers are safe by flagging five times more cases for inspection than before. At the same time it prevents overloading the safety inspection team by forwarding ten times fewer irrelevant cases.

\bibliographystyle{coling}
\bibliography{coling2020}

\end{document}